\documentclass[sigconf]{acmart}
\definecolor{yellow}{rgb}{1, 1, 0.7}
\definecolor{orange}{rgb}{1, 0.85, 0.7}
\definecolor{red}{rgb}{1, 0.7, 0.7}

\usepackage{amsfonts}
\usepackage{amsmath}
\usepackage{acronym}
\usepackage{bbm}
\usepackage{enumitem}
\usepackage{graphicx}
\usepackage{xcolor}
\usepackage{enumitem}
\usepackage{lipsum}
\usepackage{color,soul}
\usepackage{tabularx}
\usepackage{colortbl}
\usepackage[section]{placeins}
\AtBeginDocument{%
  \providecommand\BibTeX{{%
    \normalfont B\kern-0.5em{\scshape i\kern-0.25em b}\kern-0.8em\TeX}}}

\renewcommand\footnotetextcopyrightpermission[1]{} 
\begin{document}


\title{SUNDAE: \underline{S}pectrally Pr\underline{un}e\underline{d} G\underline{a}ussian Fi\underline{e}lds with Neural Compensation}

\author{Runyi Yang}
\affiliation{%
  \institution{AIR, Tsinghua University}
}
\affiliation{%
  \institution{Imperial College London}
}
\email{runyi.yang23@imperial.ac.uk}

\author{Zhenxin Zhu}
\affiliation{%
  \institution{AIR, Tsinghua University}
}
\affiliation{%
  \institution{Beihang University}
}

\author{Zhou Jiang}
\affiliation{%
  \institution{AIR, Tsinghua University}
}
\affiliation{%
  \institution{Beijing Institute of Technology}
}
\author{Baijun Ye}
\affiliation{%
  \institution{AIR, Tsinghua University}
}
\affiliation{%
  \institution{Beijing Institute of Technology}
}
\author{Xiaoxue Chen}
\affiliation{%
  \institution{AIR, Tsinghua University}
}

\author{Yifei Zhang}
\affiliation{%
  \institution{AIR, Tsinghua University}
}
\affiliation{%
  \institution{UCAS}
}

\author{Yuantao Chen}
\affiliation{%
  \institution{AIR, Tsinghua University}
}
\affiliation{%
  \institution{CUHK (SZ)}
}
\author{Jian Zhao$^\dagger$}
\affiliation{%
  \institution{EVOL Lab, Institute of AI (TeleAI), China Telecom}
}

\author{Hao Zhao$^\dagger$}
\affiliation{%
  \institution{AIR, Tsinghua University}
}
\email{zhaohao@air.tsinghua.edu.cn}

\renewcommand{\shortauthors}{R. Yang et al.}

\renewcommand{\thefootnote}{\fnsymbol{footnote}}




\begin{abstract}
\footnotetext{$^{\dagger}$Corresponding Author\vspace{-0.1cm}}
Recently, 3D Gaussian Splatting, as a novel 3D representation, has garnered attention for its fast rendering speed and high rendering quality.  However, this comes with high memory consumption, e.g., a well-trained Gaussian field may utilize three million Gaussian primitives and over 700 MB of memory. We credit this high memory footprint to the lack of consideration for the \textbf{relationship} between primitives. In this paper, we propose a memory-efficient Gaussian field named SUNDAE with spectral pruning and neural compensation. On one hand, we construct a graph on the set of Gaussian primitives to model their relationship and design a spectral down-sampling module to prune out primitives while preserving desired signals. On the other hand, to compensate for the quality loss of pruning Gaussians, we exploit a lightweight neural network head to mix splatted features, which effectively compensates for quality losses while capturing the relationship between primitives in its weights. We demonstrate the performance of SUNDAE with extensive results. For example, SUNDAE can achieve 26.80 PSNR at 145 FPS using 104 MB memory while the vanilla Gaussian splatting algorithm achieves 25.60 PSNR at 160 FPS using 523 MB memory, on the Mip-NeRF360 dataset. Codes are publicly available at \href{https://runyiyang.github.io/projects/SUNDAE/}{\textcolor{purple}{https://runyiyang.github.io/projects/SUNDAE/}}.
\end{abstract}





\begin{teaserfigure}
  \includegraphics[width=\textwidth]{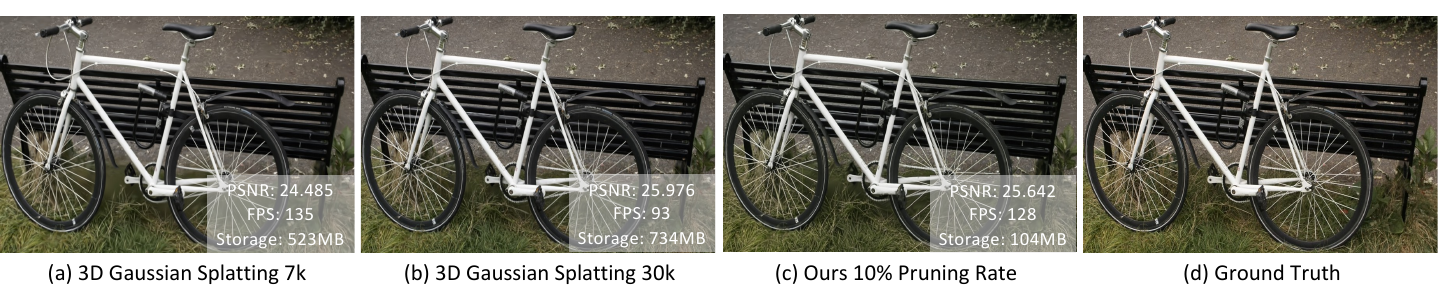}
  \vspace{-0.6cm}
  \caption{(a) 3D Gaussian splatting (3DGS) results trained for 7K iterations. (b) 3DGS results trained for 30K iterations, in which more Gaussian primitives are allocated so quality gets higher, speed gets slower, and storage gets larger compared to (a). (c) SUNDAE results by pruning 90\% primitives upon (a), being much smaller in storage, more accurate and a little bit slower than (a). Note that the storage usage is not 10\% of (a) because an additional neural compensation head is used.}
  \label{fig:teaser}
\end{teaserfigure}
\maketitle



\section{Introduction}

Representing 3D scenes has been a longstanding problem in computer vision and graphics, serving as the foundation for various VR/AR \cite{wei2024editable}\cite{peng2023synctalk} and robotics \cite{wu2023mars}\cite{zhu2023latitude}\cite{zhou2024pad} applications. With the rise of the neural radiance field (NeRF) \cite{mildenhall_nerfrepresenting_2020}, a series of methods have emerged to enhance the quality and efficiency of NeRF \cite{muller_instantneural_2022,sun2022dvgo,fridovich-keil_plenoxelsradiance_2022,chen2023mobilenerf,chen2022tensorf,xu_pointnerfpointbased_2022,yuan2023slimmerf}.  Recently, 3D Gaussian splatting (3DGS) \cite{kerbl20233d}\cite{yu2023mip}\cite{song2024sa} has been proposed as a novel 3D scene representation, utilizing a set of 3D positions, opacity, anisotropic covariance, and spherical harmonic (SH) coefficients to represent a 3D scene. Compared to neural rendering methods, this technique demonstrates notable advantages in rendering speed, rendering quality, and training time, making it widely used for applications such as 3D editing \cite{tang2023dreamgaussian, chen2023gaussianeditor} and digital twins \cite{jung2023deformable,kocabas2023hugs}.


Although 3D Gaussian splatting (3DGS) offers several advantages over other implicit 3D representations, training a 3DGS model faces a large storage challenge (i.e., checkpoint storage on ROM), as depicted in Fig.~\ref{fig:teaser} (a) and (b). This issue arises from the training process to gradually populate empty areas with Gaussian primitives, so that the rendering results can better fit input images. Compared with recent neural rendering methods \cite{muller_instantneural_2022, barron2022mipnerf360}, 3DGS requires a much larger memory cost for the same scene, which limits the application of 3GDS on mobile platforms and edge computing.

In this paper, our goal is to address the high memory cost of  3D Gaussian splatting. We introduce a memory-efficient method, which achieves low storage usage while maintaining high rendering speed and good quality. As illustrated in Fig.~\ref{fig:teaser} (c), our approach achieves photorealistic rendering quality with much lower storage.

As previously mentioned, training a vanilla 3DGS leads to a large number of Gaussian primitives, some of which are redundant.  We credit the redundancy of Gaussian primitives to the fact that these primitives are independent of each other in the 3DGS formulation. In response, we aim to improve the modeling of the \textbf{relationship} between Gaussians to reduce primitive redundancy. We achieve this through two complementary techniques: spectrally pruning the primitive graph and incorporating a neural compensation head. Consequently, our method is referred to as spectrally pruned Gaussian fields with neural compensation, abbreviated as \textbf{SUNDAE}.



\begin{figure}[htbp]
    \centering
    \includegraphics[width=0.45\textwidth]{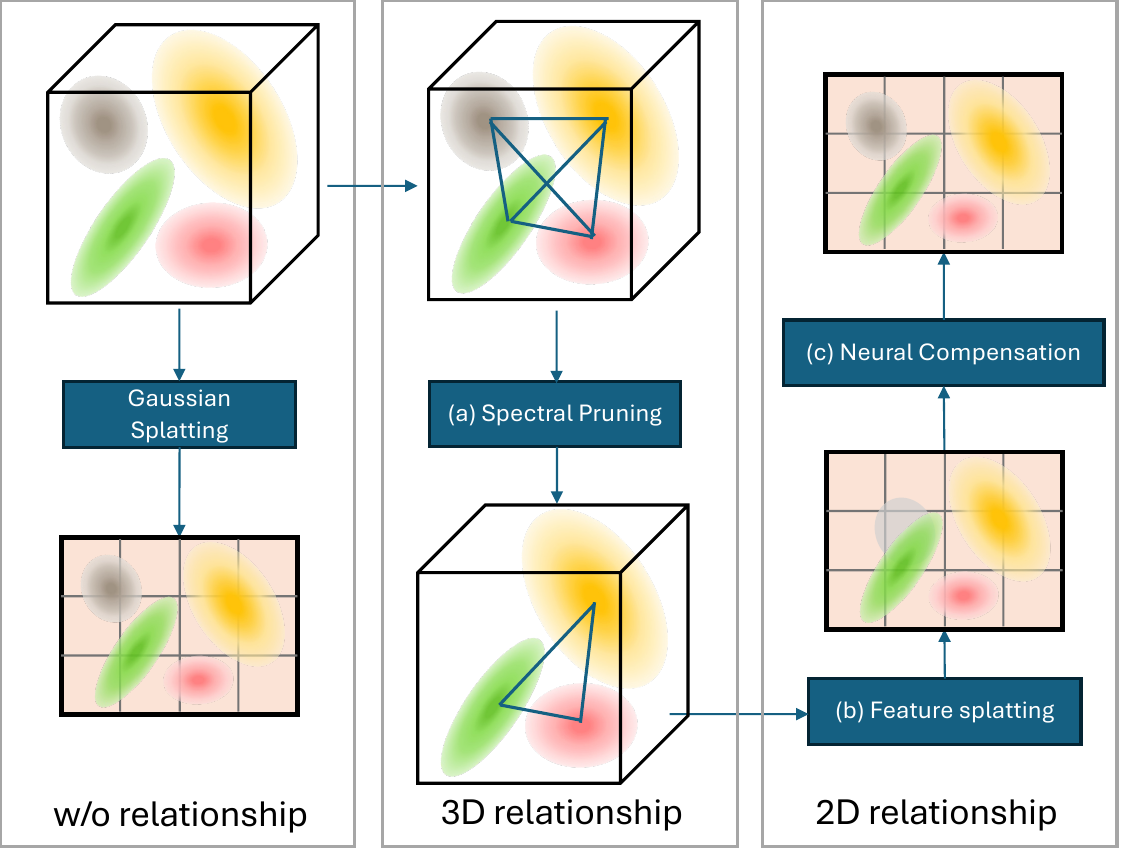}
    \caption{The left panel shows vanilla 3D Gaussian splatting, which requires a large amount of storage as it does not capture the relationship between primitives. The middle panel shows our spectral pruning technique that is based upon the relationship between 3D Gaussians. The right panel shows that the neural compensation head exploits the relationship between 2D feature splatting results to improve rendering.}
    \label{fig:relationship}
    \vspace{-1em}
\end{figure}

\textbf{Spectral Graph Pruning:} Gaussian fields utilize a collection of Gaussian primitives as the representation of the scene. As these primitives are irregularly distributed in 3D space, we propose a graph-based data structure, rather than regular structures like grids, to capture the \textbf{relationship} between these primitives. This is conceptually illustrated in the middle panel of Fig.~\ref{fig:relationship}. Specifically, We introduce the graph signal processing theory \cite{chen2015discrete} to derive an optimal stochastic sampling strategy that preserves band-limited information on this graph. By controlling the spectrum's band, we achieve flexible control over pruning ratios. For example, in Fig.\ref{fig:teaser} (c), 90\% of Gaussian primitives from Fig.\ref{fig:teaser} (a) are pruned out.

\textbf{Neural Compensation:} After spectral pruning, there is an inevitable decrease in rendering quality. To address this, we employ a neural compensation head to compensate for this quality loss, as conceptually illustrated in the right panel of Fig.~\ref{fig:relationship}. We transition from the Gaussian splatting paradigm to a feature splatting paradigm by attaching feature vectors to Gaussian primitives. Subsequently, a lightweight neural network is introduced to predict RGB values on the splatted feature images, thereby integrating information from different primitives. This allows the \textbf{relationship} between primitives to be captured in the weights of the compensation network, indirectly in the 2D domain of splatted features.

In summary, these two techniques operate in a complementary manner to tackle the absence of primitive \textbf{relationship} modeling in 3DGS. The spectral graph pruning technique removes less important primitives on the primitive graph, while the neural compensation technique integrates information from the remaining primitives. Through comprehensive qualitative and quantitative benchmarking on three datasets (Mip-NeRF360, Tanks\&Temples, and Deep Blending), we demonstrate that SUNDAE effectively reduces the size of Gaussian fields while preserving good quality and fast rendering speed. Our contributions can be summarized as:

\begin{itemize}
    \item A newly proposed primitive pruning framework for Gaussian fields based upon the spectrum of primitive graphs; 
    \item A novel feature splatting and mixing module to compensate for the performance drop caused by the pruning;
    \item State-of-the-art results, in terms of both quality and speed, on various benchmarks with low memory footprint.
\end{itemize}
\section{Related Works}
\textbf{Conventional 3D Scene Representations.}
Various 3D representations have been proposed for 3D reconstruction. Point cloud based representation \cite{fastlio2, murORB2, r3live} is extensively employed due to its simplicity and effectiveness in depicting 3D scenes. VoxelMap\cite{voxelmap} introduced a novel and efficient probabilistic adaptive voxel mapping method within the parameterized plane feature. While most of the voxel-based representation is more suitable for the downstream planning task to avoid the collision. 
Moreover, mesh-based method \cite{lorensen1998marching, ball-pivoting} 
offer distinct advantages in representing complex geometries. ElasticFusion \cite{whelan2015elasticfusion} builds on the surfel-based method to reconstruct the dense map with color. This representation is more continuous and realistic. They exploit relationship within the 3D representation in various ways, motivating our study.

\textbf{Efficient Neural Rendering.}
Vanilla NeRF~\cite{mildenhall_nerfrepresenting_2020} involves extensive neural network calculations at 3D positions on a pixel-by-pixel basis, which hinders real-time rendering. Recent approaches~\cite{muller_instantneural_2022,sun2022dvgo,fridovich-keil_plenoxelsradiance_2022,chen2023mobilenerf,chen2022tensorf,xu_pointnerfpointbased_2022,yuan2023slimmerf} have focused on improving the efficiency of Neural radiance fields, and the main perspectives include, for example, fast training, fast rendering and low memory footprint. Instant-NGP~\cite{muller_instantneural_2022} uses a multiresolution hash table to store learnable features and additionally uses a small neural network for decoding, enabling fast training. Other methods for accelerated training either use dense voxel grids, such as DVGO~\cite{sun2022dvgo} which uses a density voxel grid and a feature voxel grid for explicit and discretized volume representation, or sparse voxel grids, such as Plenoxels~\cite{fridovich-keil_plenoxelsradiance_2022} which uses a sparse 3D grid with spherical harmonics for scene representation. While these approaches significantly accelerate training, they do not render in real-time and necessitate a large consumer GPU for rendering. 3D Gaussian splatting~\cite{kerbl20233d} represents the scene using 3D Gaussians and utilizes a fast, differentiable rendering approach to achieve both high-quality novel view synthesis and real-time rendering. Our SUNDAE focuses on integrating two novel techniques to leverage the primitive relationship in Gaussian fields and improve storage efficiency. 


\textbf{Primitive-based Neural Rendering.} Different from the Vanilla NeRF that utilizes a fully implicit network for scene representation, or grid-based neural radiance fields~\cite{fridovich-keil_plenoxelsradiance_2022,muller_instantneural_2022} that evenly distribute the model parameters into each voxel grid vertex, point-based neural radiance fields~\cite{xu_pointnerfpointbased_2022,zhang_pointneuspointguided_2023,sandstrom_pointslamdense_2023a,zheng_pointavatardeformable_2023,hu_trivolpoint_2023,gao_strivecsparse_2023,huang_neuralkernel_2023,kulhanek_tetranerfrepresenting_2023,lu_urbanradiance_2023,ruckert_adopapproximate_2022,hu_point2pixphotorealistic_2023,kerbl20233d} introduce explicit graphic primitives into the construction of the neural implicit fields that allows adaptive control on the model expressiveness within limited memory consumption. Specifically, PointNeRF~\cite{xu_pointnerfpointbased_2022} associates neural features with points and performs k-NN interpolation upon arbitrary point query, allowing the network to capture finer scene details by increasing local neural point density.
Different from raytracing-based methods, ADOP~\cite{ruckert_adopapproximate_2022} and 3DGS~\cite{kerbl20233d} use rasterization techniques to render novel viewpoints. 


\textbf{Graph Signal Processing.} Graph Signal Processing aims to develop tools to process data with an irregular structure, i.e., on a graph domain. The first and most important area in graph signal processing is to design graph representations~\cite{ortega2018graph}. Groundbreaking contributions \cite{crovella2003graph} and \cite{coifman2006diffusion} presented initial samples of designs founded on the vertex and spectral domain properties, respectively. However, both frequency and time domains alone have their drawbacks \cite{ortega2018graph}. Recent research~\cite{narang2012perfect, zeng2017bipartite, ekambaram2015spline, kotzagiannidis2019splines, teke2016extending, anis2017critical, tay2015techniques} are gradually moving their focus on the development of critically sampled filterbanks that have a vertex-localized implementation as well as a spectral interpretation.  Another problem in graph signal processing is sampling graph signals. The key idea is to define a class of graph signals, i.e., bandlimited, and then define conditions to reconstruct a signal in that class. The concepts are first presented in \cite{pesenson2008sampling}. A sufficient and necessary condition for unique recovery is defined in \cite{anis2014towards} and is soon generalized to other types of graphs and signals~\cite{sandryhaila2014big, sandryhaila2014discrete, chen2015discrete}. However, sampling signals on large graphs is a great challenge due to its complexity. Some techniques require
computing the first K basis vectors\cite{chen2015discrete} and other work\cite{anis2016efficient} utilizes spectral proxies instead of exact graph frequencies to reduce complexity. Random strategy has also been proposed in \cite{puy2018random} which leads
to significantly lower complexity but less comparable performance. 

Graph signal processing has been used in a wide variety of applications, including sensor networks \cite{dong2013inference, he2016non, chen2016localization, valdivia2015wavelet}, biology networks \cite{behjat2015anatomically, atasoy2016human, leonardi2013tight}, data science \cite{huang2017collaborative, valko2014spectral, benzi2016song} and image\&pointcloud processing \cite{chen2017fast, hu2014multiresolution, tian2014chebyshev,zhang2024fastmac}. To the best of our knowledge, we are the first to introduce graph signal processing to the pruning of primitive-based neural rendering methods.
\section{Methods}


\begin{figure*}
    \centering
    \includegraphics[width=\textwidth]{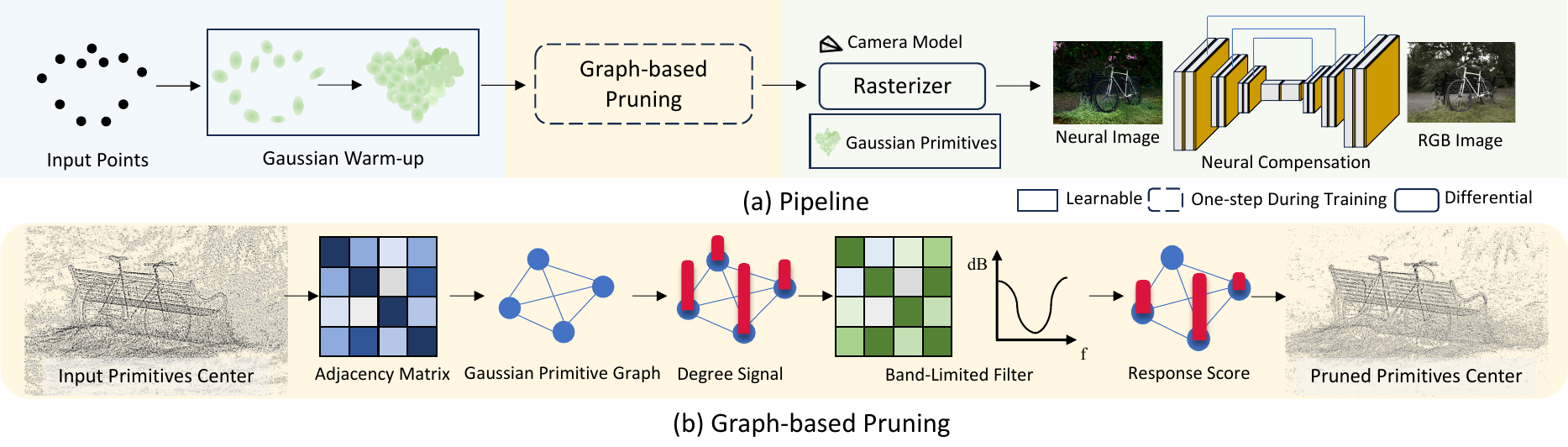}
    \caption{\textbf{(a) Pipeline}: Our proposed method warms up a 3D Gaussian field firstly, followed by a Graph-based pruning strategy to down-sample the Gaussian primitives, and a convolutional neural network to compensate the losses caused by pruning. \textbf{(b) Graph-based Pruning}: A graph based on the spatial relationship between the Gaussian primitives, is utilized for pruning post warm-up. Employing a band-limited graph filter, this process facilitates the extraction of fine details from high-frequency components, alongside capturing general features from low-frequency parts, thereby enabling a comprehensive and efficient representation of the entire scene.}
    \label{fig:main}
\end{figure*}

Given a set of images $I_i \in \mathcal{I}$, along with the corresponding camera calibration parameters $C_i$, 3DGS \cite{kerbl20233d} reconstructs the scene by representing it with Gaussian primitives yet with a high storage usage. We credit this weakness to the fact that the relationship between Guassians are not exploited in 3DGS. We introduce SUNDAE, a memory-efficient Gaussian field, featuring a graph-based pruning approach and a neural compensation module. The overall framework is depicted in Fig.~\ref{fig:main}. We begin by fitting 3D Gaussians as a ``warm-up" process in Section \ref{subsec: warmup}. To reduce memory usage, we introduce an one-step graph signal processing approach in Section \ref{subsec: graph}. This involves constructing a Gaussian primitive graph to model the relationship between primitives and utilizing a band-limit graph filter to prune redundant Gaussian primitives. To mitigate quality loss caused by pruning, we adopt a neural compensation module in Section \ref{subsec: feature render}, which restores the RGB image $\hat{I}_i$ from the neural image $F_i$ using a lightweight neural network. Finally, we introduced a continuous pruning as an alternative strategy in Sec.~\ref{sec: conti}.

 

\subsection{3D Gaussian Splatting Warm Up}
\label{subsec: warmup}

We utilize the vanilla 3D Gaussian Splatting \cite{kerbl20233d} as the initial step for generating a dense representation of a scene using Gaussian primitives. This step starts by using a sparse point cloud to initialize Gaussian centers. Then, an efficient densification strategy is exploited to increase the number of these primitives. Additionally, the rasterization process involves splatting the 3D Gaussian primitives onto the 2D plane to reconstruct input images.

\subsubsection{Gaussian Primitive Initialization}
\label{subsec: init}

The point cloud $P_c$ obtained through Structure-from-Motion (SfM) ~\cite{schonberger2016structure} serves as the initial input for representing the 3D scene using Gaussian primitives $P$. We turn the 3D coordinates $x \in P_c$ into Gaussian primitives $p \in P$, as described by the following equation:
\begin{equation}
    p(x) = \exp(-\frac{1}{2}(x)^T \Sigma^{-1}(x)),
\end{equation}

where the $\Sigma$ is defined as 3D covariance matrix in the world space. To ensure the positive semi-definiteness and to uphold the physical interpretation of the covariance matrix, an ellipsoid configuration is used to represent the 3D Gaussian covariance. The decomposition of $\Sigma$ is achieved using a scaling matrix $S$ and a rotation matrix $R$, as expressed in the equation:
\begin{equation}
    \Sigma = RSS^TR^T.
\end{equation}
This representation of anisotropic covariance is particularly advantageous for optimization processes. The generated Gaussian primitives are characterized by positions $x$, opacity $\alpha$, and a covariance matrix $\Sigma$. Additionally, the chromatic attributes of Gaussian primitives are encoded by spherical harmonic (SH) coefficients.

\subsubsection{Gaussian Primitive Densification}
\label{subsec: Gaussian}

 During the training process, all parameters of Gaussian primitives are optimized, and a densification strategy is integrated to improve representation power. Initially, Gaussian primitives exhibit a sparsity level similar to that of the point cloud generated via SfM, which is not enough for representing detailed parts of the scene, such as grass and trees in outdoor scenes. So Gaussian primitives with extensive covariance, which tend to oversimplify the geometric intricacies of detailed scene segments, are subdivided into smaller Gaussians. Meanwhile, those with minimal covariance, indicating under-representation, are duplicated to enhance coverage. We adopt the densification parameters from \cite{kerbl20233d} to increase the density of Gaussians and achieve a high-quality warm-up for the subsequent pruning process.

\subsection{Spectral Graph Pruning}
\label{subsec: graph}

After warm-up, a dense representation using Gaussian primitives incurs significant storage consumption, for example, approximately 1.33GB for the Bicycle scene in MipNeRF360 \cite{barron2022mipnerf360} dataset. We credit this inefficiency to the redundancy in primitives. However, determining which primitives are redundant is challenging without establishing \textbf{relationships} between them. Therefore, we introduce the graph signal processing theory and construct a graph based on Gaussian primitives to efficiently prune redundant primitives.

\subsubsection{Graph Signal Processing Preliminaries.}

{\textbf{Graph shift.}} We denote a weighted graph by $\mathcal{G} = (\mathcal{V}, A)$, where \(\mathcal{V}\) is the set of nodes $\{v_0,v_1,\dots,$ $ v_{N-1}\}, N=|\mathcal{V}|$ and \(A\in \mathbb{C}^{N\times N}\) is the graph shift, or the weighted adjacency matrix. Graph shift reflects the connection between the nodes using edge weight which is a quantitative description of primitive relationship. When a graph shift acts on a graph signal, it can represent the diffusion of the graph signal. In this work, we build the graph shift to model the spatial relationship among Gaussian primitives. A graph shift is usually normalized for proper scaling, ensuring \(||A||=1\). 

{\noindent \textbf{Graph signal.}} Given a graph $G=(\mathcal{V},A)$, a graph signal on this graph can be seen as a map assigning each node \(v_i\) with a value \(x_i\in \mathbb{C}\). If the order of the nodes is fixed, then the graph signal is defined as a \(N\) dimensional vector \(x=(x_1,x_2,\dots,x_N)\). In this work, the input graph node is Gaussian primitives central position $x \in \mathbb{R}^3$, the graph signal is the Euclidean distance between primitives.

{\noindent \textbf{Graph Fourier Transform.}} A Fourier transform corresponds to the expansion of a signal using a set of bases. When performing graph Fourier transform, the bases are the eigenbasis of the graph shift. For simplicity, assume that $A$ has a complete eigenbasis and the spectral decomposition of $A$ is $A = V \Lambda V^{-1}$ \cite{sandryhaila2014big}, where the eigenbasis of $A$ form the columns of matrix $V$, and $\Lambda \in \mathbb{C}^{N\times N}$ is the diagonal matrix of eigenvalues.

\subsubsection{Graph Construction}
Given a set of Gaussian Primitives $P$, we want to construct a nearest neighbor graph. The adjacent matrix $W$ of the graph is defined as:
\begin{equation}
W_{ij}=\left\{
\begin{array}{rcl}
&\exp({-\frac{||x_i- x_j||^2_2}{2*\sigma ^2}}),    \quad & { ||x_i-x_j||^2_2 < \tau} \\
 & 0,    & otherwise
\end{array}\right.
\label{eq: graph}
\end{equation}
where $x_i$ and $x_j$ are central points in $P$, $\tau$ is a hyperparameter, chosen as ten times of the minimum nearest neighborhood distance between primitives experimentally, and $\sigma$ is the variance of the distance matrix. Equation \ref{eq: graph} demonstrates that when the Euclidean distance of two Gaussian primitives is smaller than a threshold $\tau$, the two primitives are connected by the graph edge, whose weight corresponds to geometric information between two primitives in Gaussian fields. A weighted degree matrix $D$ is a diagonal matrix $D_{i,i} = \sum_j W_{i,j}$, reflecting the density around the $i$th primitive.

\subsubsection{Graph Filtering and Sampling}
\label{sec:filter mode}
We propose a band-limited graph filter, combined with a high-frequency filter and low-frequency filter (conceptually shown in Fig.~\ref{fig:main}), to catch the detailed information and general information of the scene. Specifically, the input graph signal $x$ represents the central of Gaussian primitives.

A simple design of a high-pass filter is a Haar-like one:
\begin{equation}
    \mathcal{H}_H = I - A = V (I - \Lambda) V^{-1},
\end{equation}
where $A$ is the graph shift and $V$ and $\Lambda$ are the corresponding eigenvectors and eigenvalues in diagonal form. Denote that all eigenvalues are $\lambda_i, i \in {0, 1, ..., N-1}$. We order $\lambda_i$ in descending order, thus we have $1 - \lambda_i \leq 1 - \lambda_{i+1}$ and $\lambda_0 = 1$. This indicates low-frequency response attenuates and high-frequency response amplifies \cite{chen2017fast}. Similarly, a Haar-like low-pass graph filter is:

\begin{equation}
    \mathcal{H}_L = I + \frac{A}{\lambda_0} = V (I + \Lambda / \lambda_0 ) V^{-1}.
\end{equation}

Then we have the response of the input signal $x$ corresponding to filters and the response magnitude could be written as:
\begin{equation}
    \pi_i = ||f_i||_2^2.
\end{equation}


\textbf{Controlling bandwidth with $\gamma$.} In implementation, we prune the abundant primitives according to the response magnitude of the high-pass filter. We sample total $k\%$ of all primitives, among where $\gamma$ high-frequency primitives and $(1 - \gamma)$ low-frequency primitives by querying the top $\gamma$ highest magnitude and the top $(1 - \gamma)\%$ lowest magnitude respectively. The value of $\gamma$ is ablated in Section \ref{sec: alba} and we used $\gamma = 0.5$ in main experiments to maintain consistency.

\subsection{Neural Compensation}
\label{subsec: feature render}

Although our graph-based pruning effectively removes unnecessary primitives while retaining important ones, there is inevitably a decrease in rendering quality for large pruning ratio. To address this, we employ a neural compensation network to model the relationship between primitives in the 2D domain. 

To allow neural compensation after rasterization, we need to render the 3D Gaussian primitives into neural images in a differentiable manner. Specifically, we leverage the differentiable 3D Gaussian renderer from 3DGS \cite{kerbl20233d} and switch from RGB rendering to feature rendering. The center of the Gaussian primitive is projected using the standard point rendering method:
\begin{equation}
    x_{\text{img}} = K_c ((T_c x)/(T_c x)_z),
\end{equation}
where $K_c$ and $T_c$ are the intrinsic and extrinsic parameters of camera $C$, and $x_{\text{img}}$ indicates the pixel coordinates in neural image. The covariance $\Sigma_f$ in neural image space could be formulated as:
\begin{equation}
    \Sigma_{f} = JT_c \Sigma T_c^T J^T,
\end{equation}
where the $J$ is the approximated Jacobian of the projective transformation. The neural image is computed by:
\begin{equation}
    f = \sum_{i} c_i\beta_i \prod_{j=1}^{i-1} (1-\beta_j),
\end{equation}
where $c_i$ is a feature vector instead of SHs in the original 3DGS~\cite{kerbl20233d} and $\beta$ is the result of neural image covariance $\Sigma_f$ multiplied with the opacity $\alpha$ of the Gaussian primitive.


In this manner, instead of straightforwardly rendering the RGB image like 3DGS, we obtain a neural image through the differentiable rasterizer for 3D Gaussians, which projects feature of 3D Gaussian primitives to 2D neural image $F$. Then, we utilize a lightweight neural network $\Phi$ to compensate for the quality drop post spectral pruning. This network $\Phi$ consists of a four-layer fully convolutional U-Net with skip-connections, which aggregates information from different primitives. Downsampling is performed using average pooling, and the images are upsampled using bilinear interpolation. The network takes the rasterized neural images as input and outputs the RGB images.
\begin{equation}
    \hat{I} = \Phi(F)
\end{equation}

The overall optimization is based on the difference between the rendered images and ground truth images in the dataset. The compensation network and 3D Gaussian primitives are optimized simultaneously during training. The loss function is a combination of $\mathcal{L}_1$ and a D-SSIM loss:

\begin{equation}
\begin{aligned}
    \mathcal{L} &= \lambda_1 \mathcal{L}_1 + \lambda_2 \mathcal{L}_\text{SSIM} \\
                &= \lambda_1 |\hat{I} - I| + \lambda_2 (1 - \text{SSIM}(\hat{I}, I))
\end{aligned}
\end{equation}

\subsection{Continuous Pruning as a Strategy}
\label{sec: conti}
In addition to the training-then-pruning strategy described in Sec. \ref{subsec: graph}, we further explore a strategy that integrates continuous pruning into training. Unlike training-then-pruning, which prunes out primitives from a fully densified Gaussian field, continuous pruning involves periodically removing a specific number or percentage of primitives at pre-defined intervals throughout the training process. This approach aims to consistently control the maximum number of primitives while training the 3D Gaussian field, thereby lowering \textbf{peak memory} requirements during training and allowing training on GPU devices with low GPU memory.

Empirically, the advantage of lower peak memory comes at the cost of weaker control of final memory footprint. For instance, if we prune 20\% of the primitives every 2000 iterations, the final converged state of the 3D Gaussian field might deviate from the expected 20\% reduction. Additionally, this variance can differ across different scenes, complicating the predictability and consistency of the pruning effects. Therefore, we treat the continuous pruning strategy as an alternative when needed.
\section{Experiments}
\begin{table*}[h]
\centering
\resizebox{\textwidth}{!}{
\begin{tabular}{l|l|l|l|l|l|l|l|l|l|l|l|l|l|l|l}
{Dataset} & \multicolumn{5}{c|}{{Mip-NeRF360}} & \multicolumn{5}{c|}{{Tanks\&Temples}} & \multicolumn{5}{c}{{Deep Blending}} \\

{Method/Metric} & {PSNR$\uparrow$} & {SSIM$\uparrow$} & {LPIPS$\downarrow$} & {FPS} & {Mem} & {PSNR$\uparrow$} & {SSIM$\uparrow$} & {LPIPS$\downarrow$} & {FPS} & {Mem} & {PSNR$\uparrow$} & {SSIM$\uparrow$} & {LPIPS$\downarrow$} & {FPS} & {Mem} \\
\hline
Plenoxels & 23.08 & 0.626 & 0.463 & 6.79 & 2.1GB & 21.08 & 0.719 & 0.379 & 13.0 & 2.3GB & 23.06 & 0.795 & 0.510 & 11.2 & 2.7GB \\
INGP-Base & 25.30 & 0.671 & 0.371 & 11.7 & 13MB & 21.72 & 0.723 & 0.330 & 17.1 & 13MB & 23.62 & 0.797 & 0.423 & 3.26 & 13MB \\
INGP-Big & 25.59 & 0.699 & 0.331 & 9.43 & 48MB & 21.92 & 0.745 & 0.305 & 14.4 & 48MB & 24.96 & 0.817 & 0.390 & 2.79 & 48MB \\
M-NeRF360 & \cellcolor{red}27.69 & 0.792 & 0.237 & 0.06 & 8.6MB & 22.22 & 0.759 & 0.257 & 0.14 & 8.6MB & \cellcolor{orange}29.40 & \cellcolor{orange}0.901 & \cellcolor{yellow}0.245 & 0.09 & 8.6MB \\
\hline
3DGS-7K & 25.60 & 0.770 & 0.279 & 160 & 523MB & 21.20 & 0.767 & 0.280 & 197 & 270MB & 27.78 & 0.875 & 0.317 & 172 & 386MB \\
3DGS-30K & 27.21 & \cellcolor{yellow}0.815 & \cellcolor{orange}0.214 & 134 & 734MB & \cellcolor{yellow}23.14 & \cellcolor{red}0.841 & \cellcolor{red}0.183 & 154 & 411MB & \cellcolor{red}29.41 & \cellcolor{red}0.903 & \cellcolor{orange}0.243 & 137 & 676MB \\
\hline
Ours-1\%  & 24.70 & 0.716 & 0.375 & 171 & 38MB & 20.49 & 0.703 & 0.375 & 127 & 33MB   & 26.57 & 0.861 & 0.355 & 165 & 36MB\\
Ours-10\% & 26.80 & 0.805 & 0.264 & 145 & 104MB & 22.50 & 0.787 & 0.282 & 122 & 64MB & 28.65 & 0.892 & 0.287 & 163 & 86MB\\
Ours-30\% & \cellcolor{yellow}27.24 & \cellcolor{orange}0.826 & \cellcolor{yellow}0.228 & 109 & 279MB & \cellcolor{orange}23.46 & \cellcolor{yellow}0.817 & \cellcolor{yellow}0.242 &  116 & 148MB & \cellcolor{orange}29.40 & 0.899 & 0.248 &  156 & 203MB \\
Ours-50\% & \cellcolor{orange}27.31 & \cellcolor{red}0.827 & \cellcolor{red}0.213 & 88 & 393MB  & \cellcolor{red}23.70 & \cellcolor{orange}0.830 & \cellcolor{orange}0.219 &  92 & 228MB  & 28.86 & \cellcolor{yellow}0.900 & \cellcolor{red}0.242 &   155 & 312MB \\
\end{tabular}}

\caption{Quatitative evaluation of out method with different downsampling rate compared to previous work over three datasets.}
\label{tab:method_comparison}
\end{table*}

\begin{figure*}[t]
    \centering
    \includegraphics[width=0.8\textwidth]{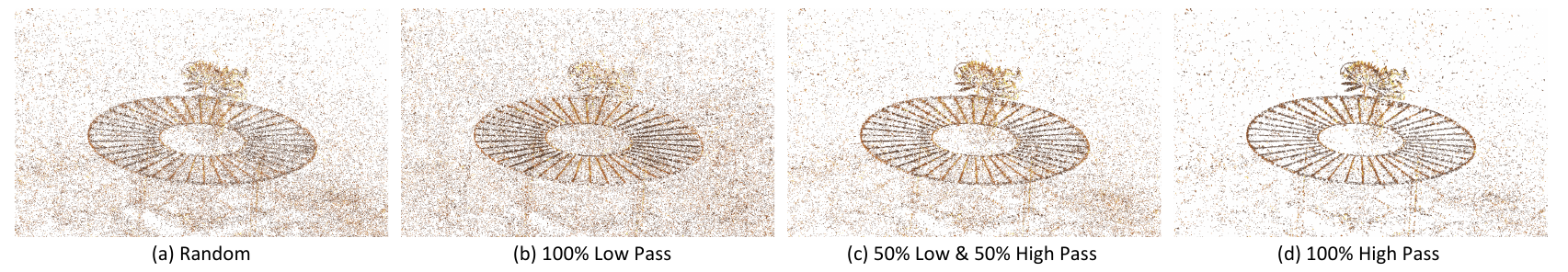}
    \vspace{-1em}
    \caption{The pruned Gaussian primitive centers with different pruning strategy.}
    \label{fig:pc_filter}

\end{figure*}

\subsection{Implemention Details}

We implement SUNDAE in Python using the PyTorch framework. The differential rasterizer component is based on the CUDA codebase, as delineated in 3DGS \cite{kerbl20233d}. Additionally, for the construction of graphs and the pruning of primitives, we employed a C++ implementation, accelerated with the Numba library for computational efficiency. All training and testing processes were conducted on a single NVIDIA RTX 3090.

Our algorithm was rigorously tested across three publicly available datasets. Specifically, we evaluated its performance on all seven scenes of the MipNeRF360 dataset, as presented in \cite{barron2022mipnerf360}. Moreover, SUNDAE was tested on two scenes from the Tanks \& Temples \cite{Knapitsch2017} and two scenes from the Deep Blending \cite{DeepBlending2018}, adhering to the benchmark criteria established in \cite{kerbl20233d}. The selected scenes encompass a diverse range of capture styles, including both bounded indoor settings and large unbounded outdoor environments.

\subsection{Quantitative Results}

\textbf{Evaluation Metrics.} In our evaluation, we employ the most widely recognized metrics in the field, PSNR, SSIM, and LPIPS. Additionally, the Frames Per Second (FPS) performance of SUNDAE was assessed by the average rendering time computed across all scenes. Memory was evaluated by calculating the average size of checkpoints in each of the scenes, which is the sum of 3D Gaussian fields and neural compensation module. We adopt the train-test split for the MipNeRF360 dataset suggested by \cite{barron2022mipnerf360}, using every 8th photo for the test, and for the rest two datasets, we followed the setting of \cite{kerbl20233d}. The tested results are shown in Tab. \ref{tab:method_comparison}. 

\begin{figure*}[htbp]
    \centering
    \includegraphics[width=0.98\textwidth]{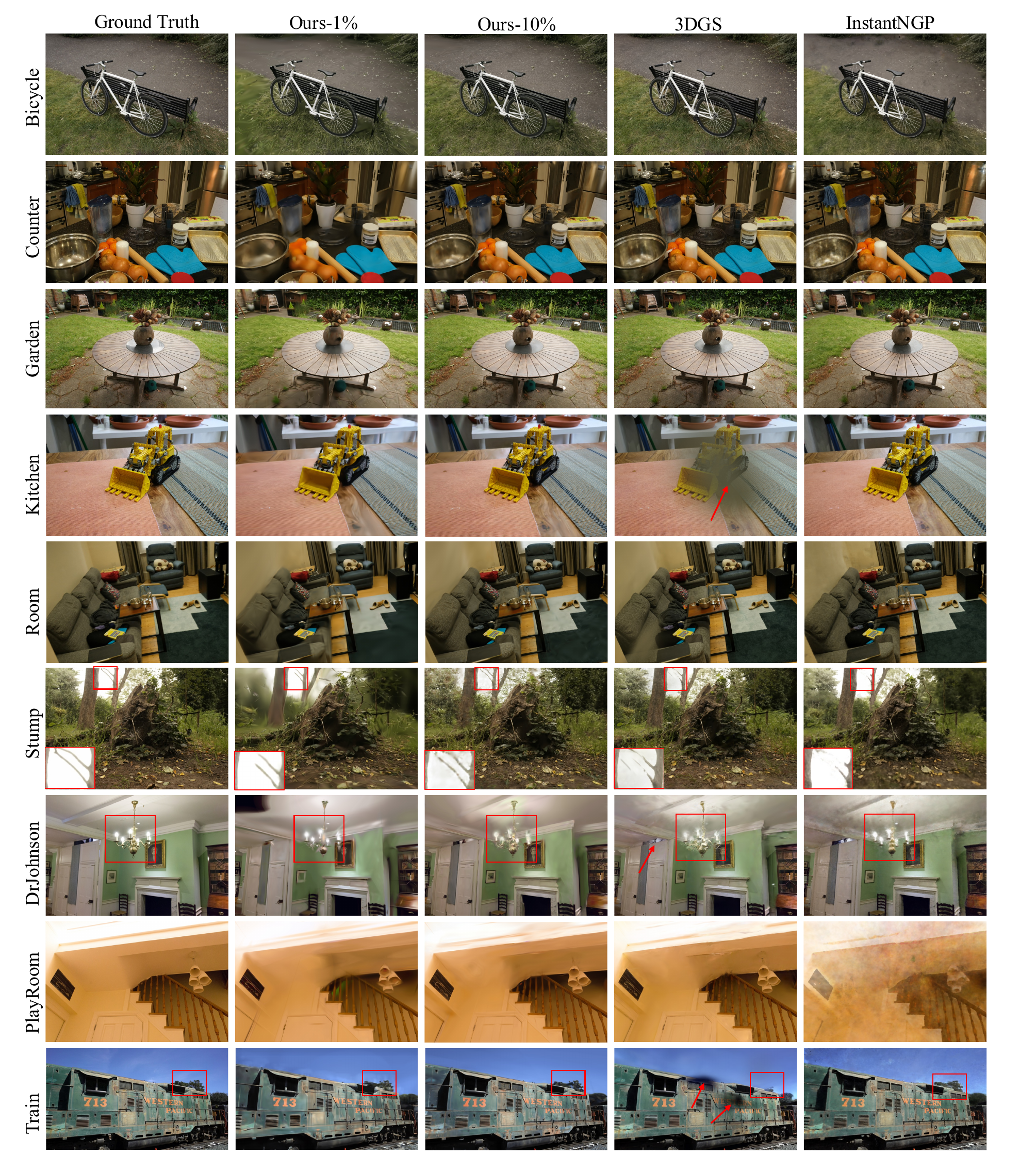}
    \caption{Qualitative results of our novel view synthesis. The scenes are, from the top
down: Bicycle, Counter, Garden, Kitchen, Room and Stump
from the Mip-NeRF360 dataset; DrJohnson, Playroom from the Deep Blending dataset and Train from Tanks\&Temples. Non-obvious differences in quality highlighted by arrows/insets.}
    \label{fig:Qualitative Results}
\end{figure*}


\textbf{Tradeoff of rendering quality, rendering time and storage.} Considering the MipNeRF360 dataset, Mip-NeRF360 method achieved the highest image fidelity with a PSNR of 27.69 and SSIM of 0.792, but had a low FPS of 0.06. Conversely, the 3DGS-7K and 3DGS-30K methods showed high FPS rates of 160 and 134, respectively, though with significant memory requirements (523MB for 7K and 734MB for 30K). Our method, particularly at 30\% and 50\% sampling rates, struck a balance between quality and efficiency, ranking within the top three in performance metrics while maintaining rapid rendering speeds and manageable memory usage (88 FPS at 393MB for 50\%). The SUNDAE variant (10\% and 1\%) also displayed remarkable efficiency, achieving a PSNR of 26.80 at 145 FPS with only 104 MB memory, and 24.70 PSNR at 171 FPS using 38 MB. This efficiency suggests that our primitive graph and neural compensation module adeptly model the relationships among primitives in 3D and 2D domains, and our pruning strategy effectively retains the primary information of the scenes. These results indicate that SUNDAE can represent scenes in a more compact manner.

As presented in Table \ref{tab:method_comparison}, our method also shows start-of-the-art performance in other datasets by using only around 50\% or even 30\% of memory. At a very low sampling rate of 1\%, our method remains competitive, closely aligning with the performance of established approaches such as Instant-NGP \cite{muller_instantneural_2022} and Plenoxels \cite{fridovich-keil_plenoxelsradiance_2022}, with minimal compromise in quality. This performance balance highlights the robustness of our spectral pruning and neural compensation techniques in managing Gaussian primitive relationships, thus there's not much decline of pruning abundant primitives.

\begin{figure}[htbp]
    \centering
    \includegraphics[width=0.45\textwidth]{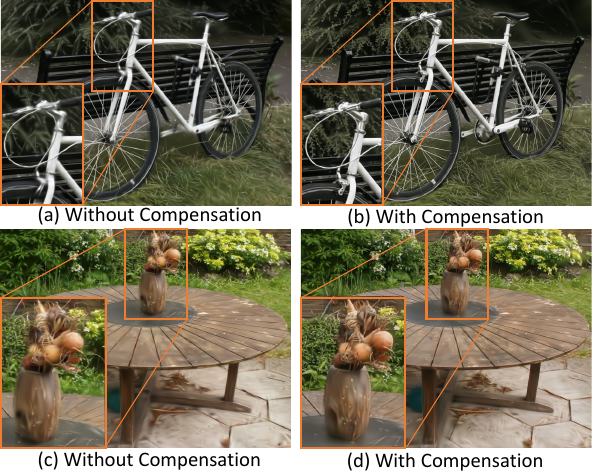}
    \caption{Visualization with and without neural compensation. }
    \label{fig:nn_comp}
    \vspace{-1em}
\end{figure}

\vspace{-1em}
\subsection{Qualitative Results}

The qualitative results can be seen in Fig. \ref{fig:Qualitative Results}. We compare the qualitative results of our SUNDAE of 1\% and 10\% sampling rate with 3DGS \cite{kerbl20233d} and InstantNGP \cite{muller_instantneural_2022}. The qualitative results show that SUNDAE could achieve a similar quality of novel view synthesis using only 10\% or even 1\% memory consumption. The graph could successfully build the relationship of primitives and the neural compensation head effectively maintains rendering quality. Interestingly, we could see from the 4-th row and the last row of Fig.~\ref{fig:Qualitative Results}, that the spectral pruning could remove the \textbf{floater} near the camera.

\textbf{Band visualization.} As seen in Fig. \ref{fig:pc_filter}, we visualize the Gaussian primitive centers post-pruning. Results show that the low-pass part could capture the background or smoothed area more, while the high-pass filter better captures the high-frequency details. The combined filter is proper to capture the high-frequency detailed object-level information as well as reserve the background points.

\subsection{Ablation Study}
\label{sec: alba}
\begin{figure}[t]
    \centering
    \includegraphics[width=0.45\textwidth]{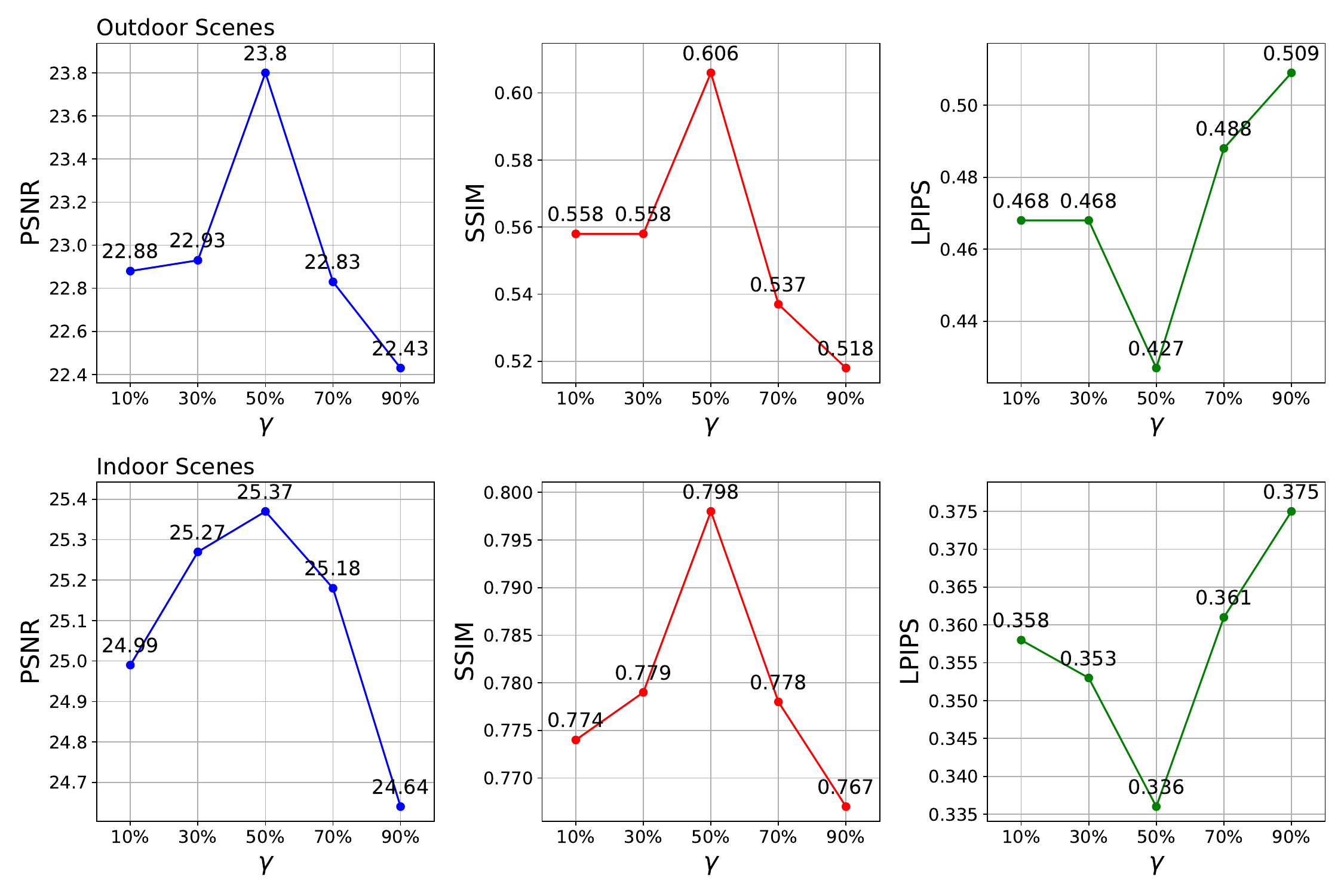}
    \vspace{-1em}
    \caption{Ablations experiment on the ratio $\gamma$ of the band-limited filter of graph based pruning. }
    \label{fig:ablation_b}

\end{figure}

\textbf{Band-limited ratio of Graph-based pruning.} The band-limited filter's ratio is represented by $\gamma$. Specifically, we sample $n$ primitives during graph-based pruning, comprising $n \times \gamma$ high-pass primitives and $n \times (1 - \gamma)$ low-pass primitives, as detailed in Sec. \ref{sec:filter mode}. As demonstrated in Figure \ref{fig:ablation_b}, we retain $1\%$ of primitives and vary the filter's ratio $\gamma$. The results indicate that $\gamma$ has a significant impact on the rendering quality. Notably, a $\gamma$ value of $50\%$ delivers the most favorable outcomes, while a disproportionate emphasis on either low-frequency or high-frequency signals leads to a deterioration in quality. This highlights the advantage of spectral pruning method as it naturally preserves important high-frequency details and low-frequency background using a 50\% ratio.



\textbf{The compensation performance of the network.} To verify the effectiveness of our neural compensation module, we conducted experiments with and without neural compensation at different sampling rates. As indicated in Table \ref{tab:ablation_nncomp}, across all sampling rates, employing neural compensation leads to improved performance compared to not using it. This is further supported by the visualization results depicted in Fig.~\ref{fig:nn_comp}. These findings demonstrate the compensatory capability of this module in mitigating the performance drops caused by spectral pruning. It is also evidenced that the relationship between primitives are well modeled.
\begin{table}[h]
\centering
\resizebox{0.5\textwidth}{!}{
\begin{tabular}{l|l|l|l|l|l|l}
{Dataset} & \multicolumn{3}{c|}{{Bicycle}} &\multicolumn{3}{c}{{Garden}}\\

{Methods} & {PSNR$\uparrow$} & {SSIM$\uparrow$} & {LPIPS$\downarrow$} & {PSNR$\uparrow$} & {SSIM$\uparrow$} & {LPIPS$\downarrow$}  \\
\hline
 1\% + neural comp & \textbf{23.23} & \textbf{0.579} &\textbf{0.449 }& \textbf{24.42} & \textbf{0.651} & \textbf{0.399} \\
 1\% w/o. neural comp & 22.79 & 0.542 & 0.466 & 24.01 & 0.614 & 0.422 \\
 \hline
 10\% + neural comp & \textbf{24.36} & \textbf{0.692} & \textbf{0.321} &\textbf{26.78} & \textbf{0.811} & \textbf{0.218} \\
 10\% w/o. neural comp & 23.93 & 0.654& 0.344 & 26.18 & 0.778 & 0.251 \\
 \hline
 30\% + neural comp & \textbf{24.40} &\textbf{0.714} &\textbf{0.271}&\textbf{27.36} & \textbf{0.843} & \textbf{0.150} \\
 30\% w/o. neural comp & 24.05 & 0.710 & 0.275 & 26.91 & 0.826 & 0.175 \\
\end{tabular}}

\caption{Ablations of neural compensation.}
\label{tab:ablation_nncomp}
\vspace{-1em}
\end{table}

\textbf{Neural Compensation Module Size.} Tab. \ref{tab:ablation_nncomp} below shows that increasing the network size does not necessarily enhance rendering quality, aligning with findings from ADOP, indicating a similar trend. We adopt 4-layer UNet of 30MB as the default setting to best balance the quality and the memory.
\begin{table}[h]
\centering
\resizebox{0.5\textwidth}{!}{
\begin{tabular}{l|l|l|l|l|l|l}
{Dataset} & \multicolumn{3}{c|}{{Bicycle}} &\multicolumn{3}{c}{{Garden}}\\

{Methods} & {PSNR$\uparrow$} & {SSIM$\uparrow$} & {LPIPS$\downarrow$} & {PSNR$\uparrow$} & {SSIM$\uparrow$} & {LPIPS$\downarrow$}  \\
\hline

 10\% w/o. neural comp & 23.93 & 0.654& 0.344 & 26.18 & 0.778 & 0.251 \\
 10\% + small NN (7.2MB)& 24.01 & 0.678 & 0.315 & 26.83 & \textbf{0.820} & 0.207\\
 10\% + regular NN (30MB) & \textbf{24.36} & \textbf{0.692} & 0.321 &\textbf{26.98} & \textbf{0.820} & 0.202 \\
 10\% + large NN (120MB) & 24.13 & 0.683 & \textbf{0.320} & 26.86 & \textbf{0.820} & \textbf{0.201}\\
\end{tabular}}

\caption{Ablations of neural compensation module size.}
\vspace{-1em}
\label{tab:ablation_nncomp}

\end{table}

\textbf{Sample More Points.} As seen in Tab. \ref{tab:method_comparison} , preserving 50\% of primitives outperformed the original 3DGS in rendering quality. We additional test keeping 80\% and keeping all primitives to test how sample rates affect the final results as seen in Tab.~\ref{tab:ablation_sampling}. Further, keeping 80\% of primitives improves rendering quality, as indicated by LPIPS, but showed minimal visual enhancement per PSNR and SSIM. Keeping all primitives (and training more epochs) could not further improve the quality, which also shows the importance of modelling the relationship of primitives. Without efficient modelling the relationship, \textbf{more primitives makes the model hard to convergence and abundant primitives are contributing negatively to the scene representation}. In addtion, our goal is to balance rendering quality with storage efficiency; however, increasing storage to 620 MB for 80\% primitives yields only slight quality improvements, diminishing storage efficiency.  
\begin{table}[h]
\centering
\resizebox{0.25\textwidth}{!}{
\begin{tabular}{l|l|l|l}

{Methods} & {PSNR} & {SSIM} & {LPIPS}  \\
\hline
3DGS & 27.21 & 0.815 & 0.214 \\ 
Ours-80\% & 27.73 & 0.835 & 0.185\\
Ours-100\% & 27.13 & 0.821  & 0.193 \\
\end{tabular}}

\caption{Ablations of sampling rate on the dataset MipNeRF360.}

\label{tab:ablation_sampling}
\vspace{-2em}
\end{table}

\textbf{Continuous Pruning.} In Sec.~\ref{sec: conti}, we propose a continuous pruning strategy. We tested it on Bicycle and Counter scenes in MipNeRF360 dataset according to different pruning interval iterations and pruning rate. As seen in Tab.~\ref{tab:pruning}, points are the number of primitives after training, and ratio is the rough ratio of number of primitives and the original 3DGS after training. Results show that this strategy could reduce the peak memory, but it is hard to control the final memory (reflected by points and ratio). And it would cause a quality loss for its parameter sensitivity. So we justify our training-then-pruning strategy but still provide the continuous pruning strategy as an alternative in our open-source toolbox.
\begin{table}[htbp]
  \centering
  \caption{Evaluation for different pruning strategies on MipNeRF360 dataset.}
  \label{tab:pruning}
  \resizebox{0.5\textwidth}{!}{
  \begin{tabular}{@{}lccccccc@{}}
    \toprule
    Pruning Strategy & Scene& PSNR & SSIM & LPIPS & Points & Ratio & Peak Memory\\ 
    \midrule
    Prune every 2k [70\%] & Bicycle & 24.50 & 0.730 & 0.230 & 531w &~70\% &7032MB\\
    Prune every 3k [70\%] & Bicycle & 24.47 & 0.730 & 0.224 & 673w &~80\% &8301MB \\
    \midrule
    Training-then-pruning [50\%] & Bicycle & 24.46 & 0.719 & 0.251 & 421w &~50\% & 18036MB\\
    Prune every 2k [50\%] & Bicycle & 24.35 & 0.723 & 0.243 & 412w &~50\% & 6386MB\\
    Prune every 2k [30\%] & Bicycle & 24.17 & 0.699 & 0.264 & 272w &~30\% & 6338MB\\
    Prune every 2k [10\%] & Bicycle & 23.52 & 0.622 & 0.363 & 156w &~20\% & 5056MB\\
    \midrule
    Training-then-pruning [50\%] & Counter & 28.10 & 0.882 & 0.192 & 63.4w &~50\% & 3730MB \\
    Prune every 2k [30\%] & Counter & 27.56 & 0.863 & 0.160 & 62.2w &~50\% & 1978MB\\
    Prune every 2k [10\%] & Counter & 27.00 & 0.838 & 0.194 & 42.1w &~30\% & 1805MB\\
    Prune every 2k [1\%] & Counter & 24.54 & 0.777 & 0.194 & 15.7w &~10\% & 1542MB\\
    \bottomrule
  \end{tabular}}
\end{table}

\subsection{Efficiency Evaluation}
For details on training time, CUDA memory, rendering FPS, and ROM storage, refer to Tab. \ref{tab:train}. Notably, `Ours-50\%' achieves state-of-the-art rendering quality with an acceptable training duration of 1.41 hours, while achieving real-time rendering, and significantly lowering both CUDA memory usage during training and ROM storage.
\begin{table}[h]
\centering
\resizebox{0.5\textwidth}{!}{
\begin{tabular}{l|c|c|c|c}
{Methods} & {Training Time} & {FPS} & {CUDA Mem} & {ROM Storage}  \\
\hline
3DGS      & 37.98min & 134 & 10.53 GB& 1.33 GB \\
Ours-1\%  & 51.99min & 171 & 2451 MB& 43.6 MB \\ 
Ours-10\% & 1.10h & 145 & 3169 MB& 141 MB \\
Ours-30\% & 1.26h & 109 & 4790 MB& 447 MB\\
Ours-50\% & 1.41h & 88  & 6426 MB& 710 MB\\
\end{tabular}}

\caption{Evaluation metrics on the bicycle scene of  MipNeRF360 dataset.}
\label{tab:train}

\end{table}

\vspace{-2em}
\section{Conclusion}

In this paper, we present a novel method of spectrally pruned Gaussian fields (SUNDAE) with neural compensation, that robustly and efficiently models the relationship between Gaussian primitives by introducing the graph signal processing framework and mix the information from different primitives to compensate the information loss caused by pruning. We use the spatial information between Gaussian primitives to construct a graph to model the relationship and spectrally prune the less important ones. A lightweight neural network is utilized to compensate for the inevitable rendering quality loss post-pruning. Experimental results show that SUNDAE well maintains the efficency of 3DGS while being much more smaller on a wide spectrum. 

{
\bibliographystyle{ACM-Reference-Format}
\bibliography{sample-base}
}

\end{document}